\documentclass{article}

\usepackage[main,final]{iaseai26}

\usepackage[utf8]{inputenc}
\usepackage[T1]{fontenc}
\usepackage{booktabs}
\usepackage{amsmath, amssymb, amsthm}
\usepackage{graphicx}
\graphicspath{{../}}
\usepackage{hyperref}
\usepackage{siunitx}
\usepackage{longtable}
\usepackage{array}
\usepackage{csquotes}
\usepackage{tabularx}
\usepackage{xcolor}
\usepackage{multirow}

\title{Measuring What Matters: The AI Pluralism Index}

\author{Rashid Mushkani\\
Universit\'e de Montr\'eal\\
Mila -- Qu\'ebec AI Institute\\
\texttt{rashidmushkani@gmail.com}}

\begin{document}

\maketitle

\begin{abstract}
Artificial intelligence systems increasingly mediate knowledge, communication, and decision making. Development and governance remain concentrated within a small set of firms and states, raising concerns that technologies may encode narrow interests and limit public agency. Capability benchmarks for language, vision, and coding are common, yet public, auditable measures of pluralistic governance are rare. We define \emph{AI pluralism} as the degree to which affected stakeholders can shape objectives, data practices, safeguards, and deployment. We present the AI Pluralism Index (AIPI), a transparent, evidence‑based instrument that evaluates producers and system families across four pillars: \emph{participatory governance}, \emph{inclusivity \& diversity}, \emph{transparency}, and \emph{accountability}. AIPI codes verifiable practices from public artifacts and independent evaluations, explicitly handling \emph{Unknown} evidence to report both lower‑bound (“evidence”) and known‑only scores with coverage. We formalize the measurement model; implement a reproducible pipeline that integrates structured web and repository analysis, external assessments, and expert interviews; and assess reliability with inter‑rater agreement, coverage reporting, cross‑index correlations, and sensitivity analysis. The protocol, codebook, scoring scripts, and evidence graph are maintained openly with versioned releases and a public adjudication process. We report pilot provider results and situate AIPI relative to adjacent transparency, safety, and governance frameworks. The index aims to steer incentives toward pluralistic practice and to equip policymakers, procurers, and the public with comparable evidence.
\end{abstract}

Index: \url{https://aipluralism.wiki/}\\
Code: \url{https://github.com/rsdmu/aipi-pluralism-index}

\section{Introduction}
Artificial intelligence is now infrastructural in search, education, media, health, finance, and public administration. Decisions about data curation, model objectives, deployment defaults, safety mitigations, and redress are largely made by a small number of labs and government programs. International guidance warns that such concentration undermines democratic legitimacy and risks sustained exclusion of those most affected \citep{unesco2021,Crawford2021,benkler2006wealth,coe2024convention}. Users face a second challenge: for many everyday tasks, top systems often produce broadly similar outputs; headline capability differences are therefore less informative for responsible selection \citep{RAY2023,BommasaniEtAl2021,chiang2024arena,helm2025capabilities}. When functionality converges, the governance properties of systems become a salient differentiator. 

Existing evaluations compare model capabilities and user preferences \citep{RAY2023,chiang2024arena,stanford_helm}. Other efforts synthesize ethics and disclosure guidance \citep{Jobin2019,Hagendorff2020,mushkani2025_rights,bommasani2023fmti,techpolicy2024_fmti} and safety governance \citep{bengio2024safety,fli_asi2025,eu_ai_act,iso42001}. Regulatory frameworks articulate process expectations—risk management, incident handling, documentation, accessibility—but do not themselves yield comparable scores across providers \citep{Tabassi2023,euai2024}. There is no comprehensive, public, auditable instrument that centers pluralism—whether affected stakeholders can influence objectives, safeguards, and deployment—and measures participation, inclusion, transparency, and accountability in verifiable ways. This paper addresses that gap.

\paragraph{Contributions.}
This paper introduces the AI Pluralism Index (AIPI), a community‑governed protocol that: (i) formalizes an indicator hierarchy operationalizing pluralism across four pillars grounded in international norms and empirical work \citep{Jobin2019,Arnstein1969,Bohman2000,Mitchell2019,gebru2021,Goggin2019,unesco2021,oecd2019ai,fung2006,wcag22}; (ii) implements a transparent evidence pipeline and scoring with two principled treatments of \emph{Unknown}; (iii) evaluates reliability and construct validity via inter‑rater agreement, cross‑index correlations, and sensitivity analysis; and (iv) publishes system‑ and provider‑level results with evidence provenance through a versioned, open governance process aligned with standards for incident reporting and vulnerability disclosure \citep{iso29147,iso30111,ai_incident_db}.

\section{Related work}
\label{sec:related}
Capability benchmarks such as Chatbot Arena and HELM provide large‑scale comparative results for tasks under controlled protocols \citep{RAY2023,chiang2024arena,stanford_helm}. These efforts inform model selection for performance but are largely orthogonal to governance. Transparency syntheses assess disclosure practices and ethics guidance \citep{Jobin2019,Hagendorff2020,bommasani2023fmti,techpolicy2024_fmti}. Safety indices and governance assessments focus on risk management commitments and safeguards \citep{bengio2024safety,fli_asi2025}. Regulatory instruments set expectations for trustworthy AI without creating measurement tools: the EU AI Act defines risk‑based obligations, documentation, and oversight \citep{euai2024}. The NIST AI Risk Management Framework provides organizational process scaffolding \citep{Tabassi2023,iso42001}. AIPI complements these efforts by offering an auditable, evidence‑driven index focused on pluralism: participation, inclusion, transparency, and accountability practices that shape how systems are built and governed.

\section{Concept and scope}
\label{sec:concept}
\begin{figure}[t]
  \centering
  \includegraphics[width=\linewidth]{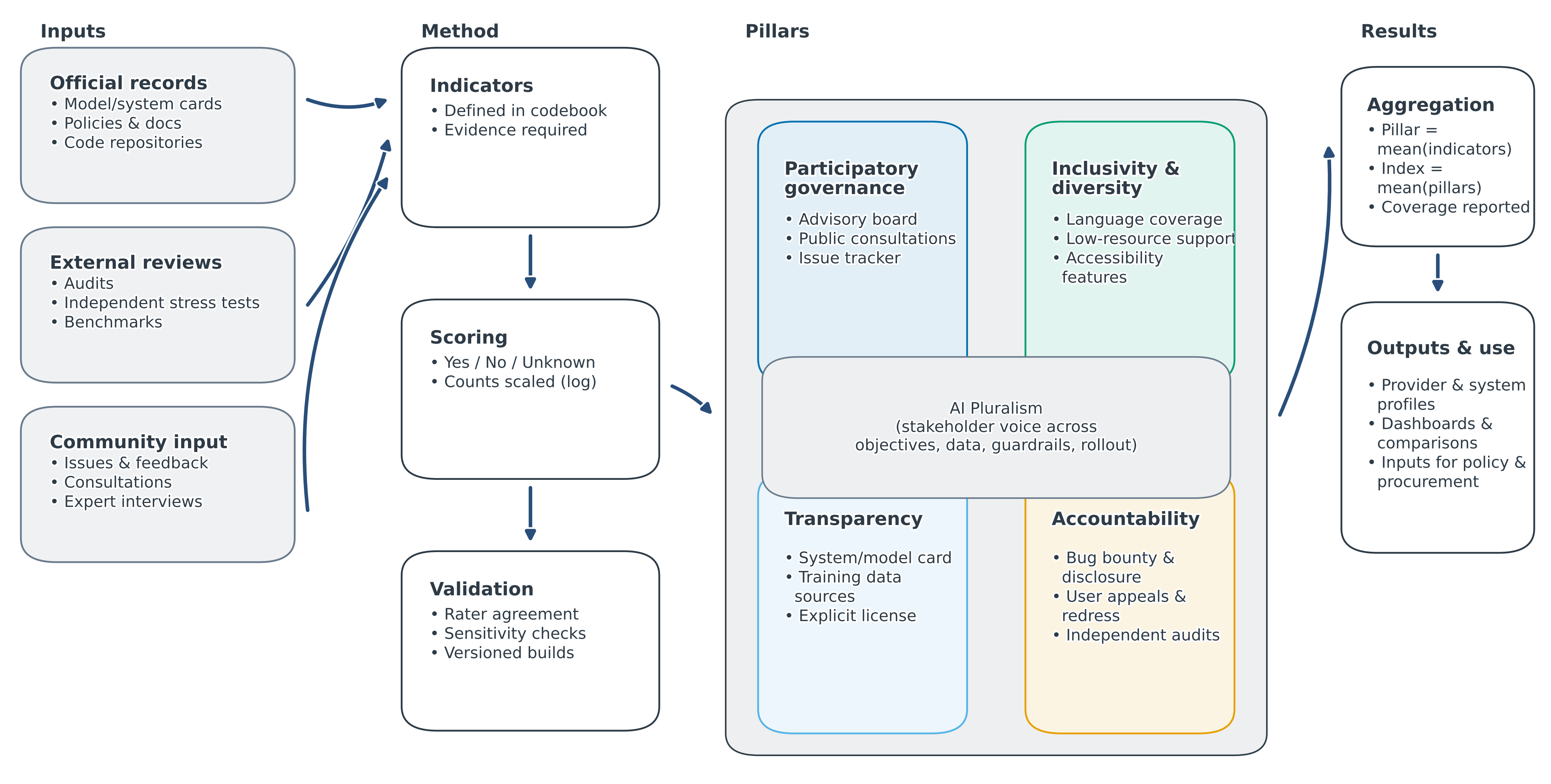}
  \caption{Conceptual framework of the AI Pluralism Index (AIPI). Verifiable evidence is coded into indicators, aggregated into four equal‑weighted pillars (Participatory governance, Inclusivity \& diversity, Transparency, and Accountability), and averaged to produce AIPI. Open governance, versioned releases, and validation provide auditability.}
  \label{fig:aipi_framework}
\end{figure}

We define \textbf{AI pluralism} as a property of socio‑technical development and deployment processes whereby affected stakeholders can meaningfully shape objectives, constraints, and outcomes. Pluralism concerns both governance and production. The index is anchored in four equal‑weighted pillars (Fig.~\ref{fig:aipi_framework}): participatory governance, inclusivity and diversity, transparency, and accountability. These pillars align with traditions in participation studies and human–computer interaction—Arnstein’s ladder and deliberative traditions emphasize degrees of influence and institutional design \citep{Arnstein1969,Bohman2000,fung2006}; Value Sensitive Design operationalizes stakeholder engagement and value articulation in technology development \citep{Umbrello2021,vsd2019}. Inclusivity requires attention to representation in teams, language access, and accessibility \citep{Goggin2019,wcag22}. Transparency and accountability cover documentation, incident reporting, audits, redress, and vulnerability handling \citep{Mitchell2019,gebru2021,Lepri2018,iso29147,iso30111}. The unit of analysis is the \emph{producer} (company, lab, consortium, or public agency), with optional system‑family granularity when practices differ across offerings. The scope is global and normative, aligning with recognized principles rather than a single jurisdiction. Because AIPI is intended as a global index, participatory governance indicators distinguish between geographically bounded engagement (e.g., only in a producer’s home jurisdiction) and engagement that is demonstrably multi-region. We do not apply separate region or population weights in aggregation; instead, geographic scope is captured within relevant indicator rubrics and reported as metadata in the evidence graph.

\section{Design choices: indicators, observability, and consequences}
\label{sec:design-choices}
AIPI credits only practices that leave public, auditable traces. Indicators must be \emph{observable} through public artifacts or external evaluations, \emph{verifiable} via durable references and timestamps (with archives when permitted), and \emph{consequential} in the sense that they allocate decision rights, impose process requirements, or document changes. This approach reduces coder discretion, enables third‑party audit, and aligns scoring with practices that affected stakeholders can inspect and use. More broadly, AIPI follows established methodology for constructing composite indicators, including explicit conceptual framing, transparent normalization and weighting choices, and sensitivity analysis \citep{oecd2008composite}.

Observability requires a primary artifact rather than a claim about one. Acceptable artifacts include policies, model or system cards, release notes, governance records (e.g., minutes, charters, response summaries), audit reports, and entries in recognized registries. To receive full credit, an artifact must identify the decision locus (who decides and on what scope), describe the process or scope (e.g., advisory remit, audit coverage, data governance components), and record outcomes or dispositions (e.g., decisions, remediation, linked change logs). A model card that specifies purpose, training data provenance, intended uses, limitations, subgroup performance, and update history is observable evidence \citep{Mitchell2019,gebru2021}. An independent audit that publishes scope, method, findings, and remediation is observable evidence \citep{Lepri2018,raji2020closing}. Marketing pages or press releases without scope, process, or outcomes are not.

Verifiability requires that evidence be durable and comparable across releases. Each artifact must have a persistent URL and a clear date; where possible we store or reference an archived copy. Versioned artifacts are coded by version. Materials behind paywalls, NDAs, or logins do not qualify as primary evidence because affected stakeholders cannot inspect them. When a standards claim is made (e.g., “WCAG 2.2 AA”), the artifact should state scope (surfaces, products, versions) and date; claims without scope or date receive partial credit under the ordinal rubric in Table~\ref{tab:scoring-rubric} \citep{Goggin2019,wcag22}.

Consequence is assessed by whether a practice changes procedures, authority, or obligations. In participatory governance, credit attaches to mechanisms that record input and document dispositions, such as public consultations with response summaries, advisory bodies with published remits and minutes, and external community red teaming with published scope and outcomes \citep{Arnstein1969,fung2006}. Generic feedback channels without dispositions do not receive credit.

For inclusivity and diversity, we count language support only when multiple user‑facing components are present for the flagship surface. A language is counted if at least three of the following are documented: localized user interface, localized help or FAQ, localized policies and safety disclosures, and a localized redress or appeals channel. Auto‑translation or preview‑only modes are not counted. Accessibility claims are assessed with stated scope and date; self‑declarations without scope receive partial credit; independent verification receives full credit \citep{Goggin2019,wcag22}. Partnerships outside high‑income markets count when they involve co‑production (e.g., co‑authored resources, shared maintenance) rather than one‑off donations. Statements of commitment without accompanying statistics, methods, or changes do not receive credit. For multilingual evaluation and collaboration in lower‑resourced settings, we draw on evidence from participatory language work \citep{Nekoto2020}.

Transparency indicators credit documentation that is decision‑relevant for users, procurers, and auditors. Model and system cards are coded when they cover purpose, data provenance, intended uses, limitations, subgroup results, and update history \citep{Mitchell2019,gebru2021}. Release notes that link governance or safeguard changes to dates and versions are counted. Narrative blog posts without an associated artifact are not counted.

Accountability indicators distinguish policy statements from implemented processes. Vulnerability disclosure is credited when inbound channels, scope, and safe‑harbor language are documented; vulnerability handling is credited when roles, triage, and timelines are documented, with references to recognized practice \citep{Lepri2018,Leslie2019}. Independent audits with public findings and remediation timelines are credited above self‑attestations. Incident postmortems are credited when they supply technical and procedural details; contributions to public learning repositories and impact assessment practices are counted as additional evidence \citep{Reisman2018,ai_incident_db}.

Where mature standards exist, coding is anchored to them to reduce ambiguity across providers and releases. For accessibility we assess the presence of concrete conformance claims and independent verification \citep{Goggin2019}; for security disclosure and handling we look for documented processes \citep{Lepri2018,Leslie2019}. Standards references without scope or date do not suffice; conformance statements with scope and date receive partial credit; verification (e.g., audit or test report) receives full credit \citep{wcag22,iso29147,iso30111}.

Counting indicators (e.g., number of supported languages, consultations, or disclosed incidents) follow inclusion rules designed to limit inflation. Counts are mapped using the tempered logarithmic transform in Eq.~(1) to cap marginal influence and reduce sensitivity to outliers (Sec.~\ref{sec:model}). Disclosed incidents are not penalized; when paired with postmortems they are treated as accountability evidence.

Boundary cases follow consistent rules. Primary artifacts take precedence over secondary reporting; without a primary artifact, no credit is assigned. Provider‑level artifacts apply to all systems only when scope is explicit; otherwise credit is limited to the named systems. Artifacts predating the evidence window are coded but may receive partial credit for staleness. Gated materials do not qualify as observable. When public claims conflict, we code to the more conservative interpretation and record the discrepancy in adjudication logs.

Certain items are excluded by design: principle statements without mechanisms; advisory bodies without a charter, roster, cadence, and outputs; redress channels that are indistinguishable from generic support (no service levels, no appeal paths); language claims based only on translation without localized policies or redress; and private assurances made to the project team.

The scoring implications are direct. In the evidence model, missing public artifacts are treated as zero, which assigns credit only when documentation exists. The known‑only model averages over observed indicators. Coverage is reported to separate documentation gaps from practice and to signal where rankings may be sensitive to missing evidence (Sec.~\ref{sec:model}).

\section{Measurement model}
\label{sec:model}
\begin{figure}[t]
  \centering
  \includegraphics[width=\linewidth]{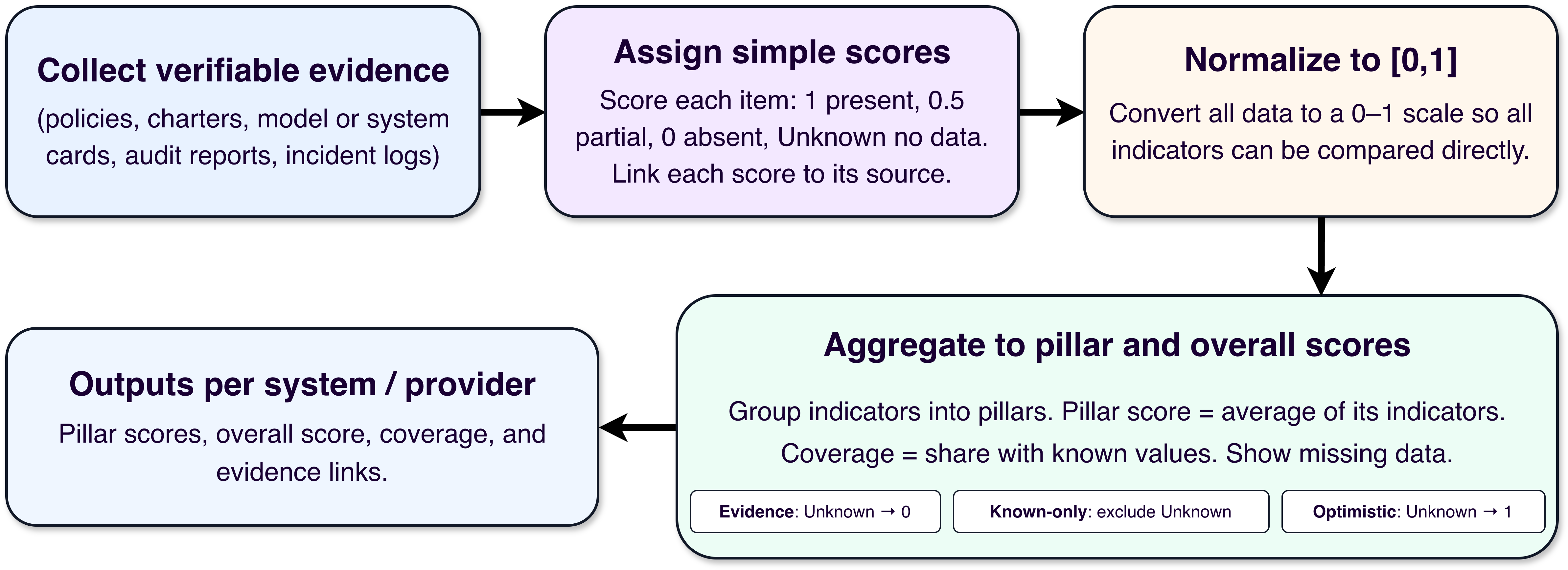}
  \caption{Simple schematic of the AIPI measurement model. Indicators are coded from verifiable artifacts, normalized, aggregated to pillar means, and reported as an interval via explicit treatments of \emph{Unknown} (evidence, known-only, optimistic) with coverage.}
  \label{fig:measurement_schematic}
\end{figure}

\begin{figure}[t]
  \centering
  \includegraphics[width=\linewidth]{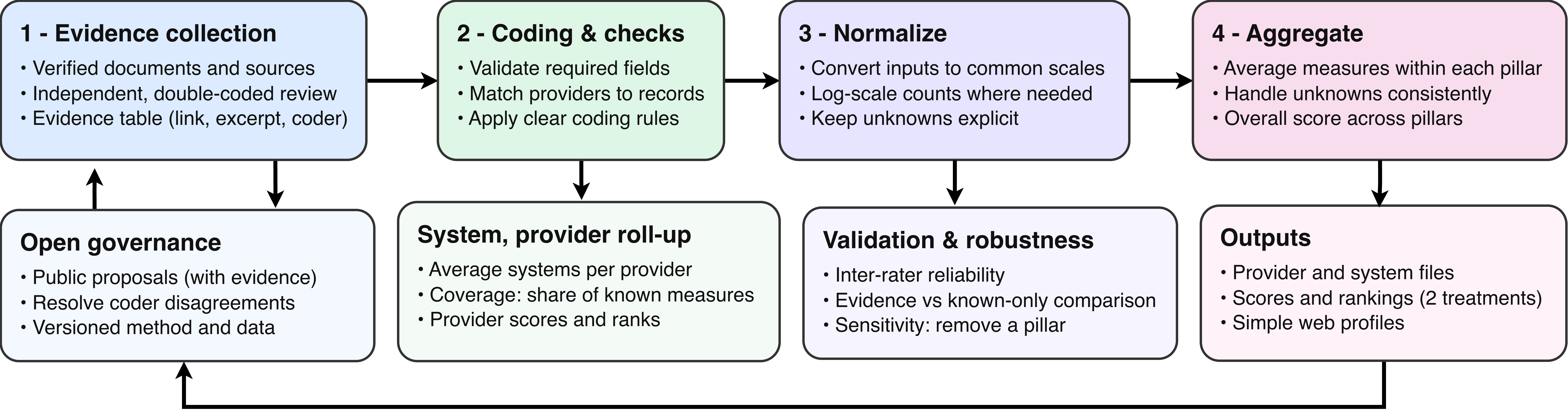}
  \caption{AIPI methodology. Verifiable artifacts, structured web \& repository analysis, external evaluations, and expert interviews feed a coded dataset with evidence links. After validation and normalization, pillar scores are aggregated under two treatments of Unknown (\emph{evidence} vs.\ \emph{known‑only}) to produce system‑ and provider‑level AIPI. Validation reports inter‑rater reliability, cross‑index correlations, and sensitivity.}
  \label{fig:method}
\end{figure}

\begin{figure}[t]
  \centering
  \includegraphics[width=\linewidth]{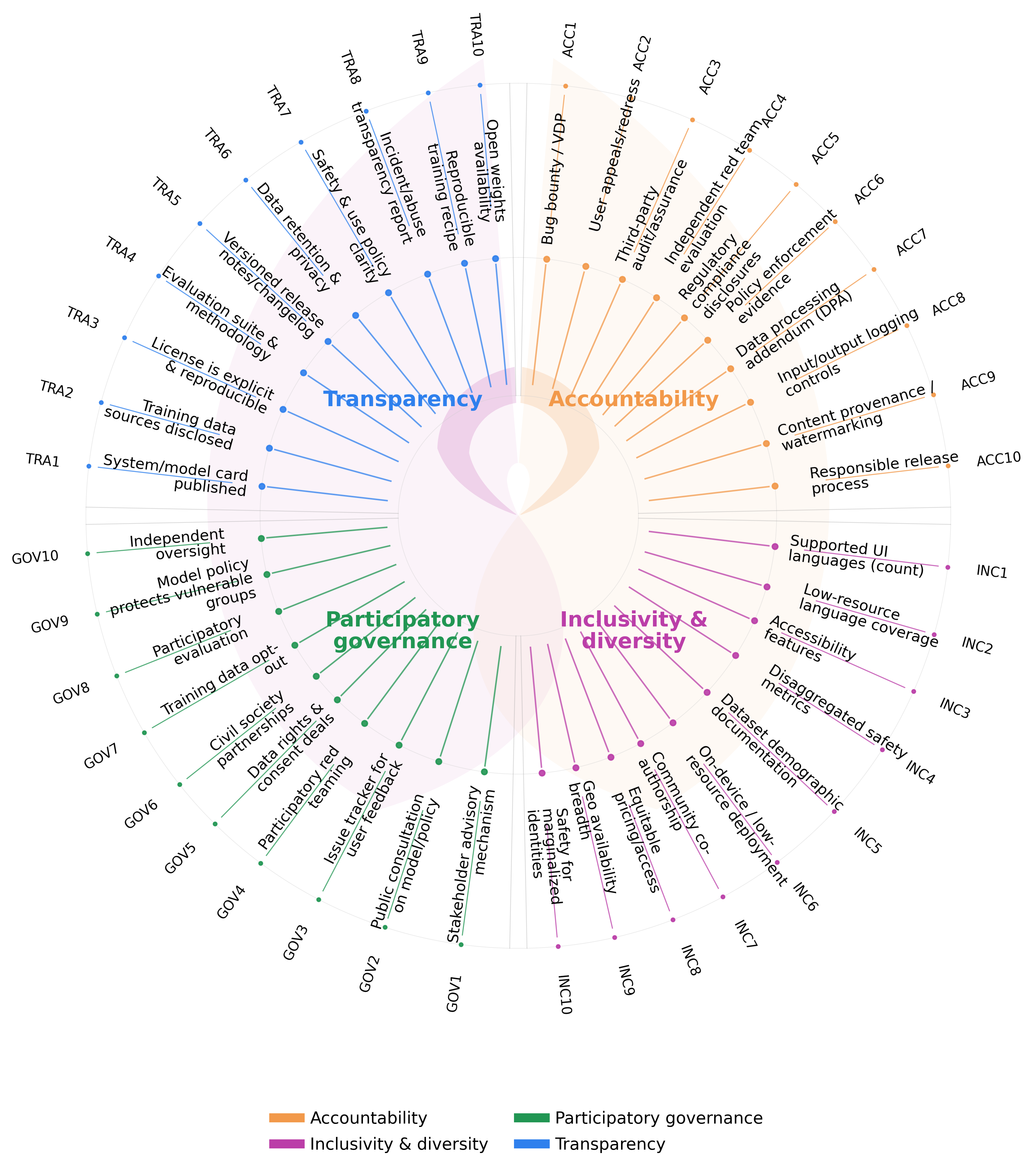}
  \caption{Pillar–subpillar–indicator hierarchy. The radial diagram enumerates indicators within each pillar and subpillar; the outer band lists indicator IDs.}
  \label{fig:aipi_02}
\end{figure}

AIPI aggregates evidence-coded indicators into pillar and overall scores with explicit treatment of \emph{Unknown}. Figure~\ref{fig:measurement_schematic} provides a visual overview of the scoring logic presented in this section. Figure~\ref{fig:method} summarizes the overall pipeline, and Figure~\ref{fig:aipi_02} presents the pillar–subpillar–indicator hierarchy.

\subsection{Indicator normalization}
Indicators are coded from verifiable artifacts or external evaluations and mapped to $[0,1]$.
Binary indicators map $\text{Yes}\mapsto 1$, $\text{No}\mapsto 0$, and \emph{Unknown} to missing. Ordinal indicators use a generic rubric (Table~\ref{tab:scoring-rubric}) mapping $\{0,1,2\}\mapsto\{0,0.5,1\}$. For counts $c$ we use a tempered logarithmic mapping
\begin{equation}
  s(c)=\min\!\left(1,\frac{\log(1+c)}{\log\!\left(1+c_{\mathrm{ref}}\right)}\right),
\end{equation}
where $c_{\mathrm{ref}}$ is the $95^\mathrm{th}$ percentile within the release. The transform is monotone, bounded, less sensitive to outliers than linear scaling, and comparable across releases because $c_{\mathrm{ref}}$ is frozen per version.

\subsection{Aggregation and properties}
Let $i$ index systems (or producers) and $p \in \{\mathrm{PG},\mathrm{ID},\mathrm{TR},\mathrm{AC}\}$ index pillars. For indicator set $\mathcal{J}_p$, the \emph{known‑only} pillar score averages normalized values $s_{ijp}$ over indicators with evidence:
\begin{equation}
    S^{\mathrm{known}}_{ip} = \frac{1}{|\mathcal{J}^{\mathrm{known}}_{ip}|} \sum_{j \in \mathcal{J}^{\mathrm{known}}_{ip}} s_{ijp}.
\end{equation}
The \emph{evidence} model treats Unknown as $0$:
\begin{equation}
  S^{\mathrm{evid}}_{ip} = \frac{1}{|\mathcal{J}_{p}|} \sum_{j \in \mathcal{J}_{p}} s^{*}_{ijp},
  \qquad
  s^{*}_{ijp}=
  \begin{cases}
    s_{ijp} & \text{if known},\\
    0 & \text{if Unknown}.
  \end{cases}
\end{equation}
Coverage is reported as $C_{ip} = |\mathcal{J}^{\mathrm{known}}_{ip}| / |\mathcal{J}_p|$. Overall scores average the pillars:
\begin{equation}
  \mathrm{AIPI}^{\bullet}_{i} = \frac{1}{4}\sum_{p} S^{\bullet}_{ip}, \qquad \bullet \in \{\mathrm{known},\mathrm{evid}\}.
\end{equation}
A provider’s score averages across its $K$ systems:
\begin{equation}
    \mathrm{AIPI}^{\bullet}_{\text{provider}} = \frac{1}{K} \sum_{k=1}^{K} \mathrm{AIPI}^{\bullet}_{i_k}.
\end{equation}

\paragraph{Bounding and reporting.}
An upper bound treats Unknown as $1$:
\begin{equation}
S^{\mathrm{opt}}_{ip}=\frac{1}{|\mathcal{J}_{p}|}\sum_{j\in\mathcal{J}_p} s^{+}_{ijp},
\qquad
s^{+}_{ijp}=
\begin{cases}
  s_{ijp} & \text{if known},\\
  1 & \text{if Unknown}.
\end{cases}
\end{equation}
For every system and pillar:
$S^{\mathrm{evid}}_{ip} \le S^{\mathrm{known}}_{ip} \le S^{\mathrm{opt}}_{ip}$. The reporting convention is to publish $(\mathrm{AIPI}_{\text{evid}}, \mathrm{AIPI}_{\text{known}}, \mathrm{AIPI}_{\text{opt}})$ as a closed interval with a marked midpoint, plus pillar‑level coverage intervals $\left(C_{ip}^{\min}, C_{ip}^{\max}\right)$ where the minimum treats un‑attributable evidence as uncovered and the maximum includes neutral evidence verified by third parties (Sec.~\ref{sec:future}).

\paragraph{Estimator guidance.}
For procurement and safety‑critical use, we recommend $\mathrm{AIPI}_{\text{evid}}$ (the lower‑bound score) with minimum floors for the overall score, each pillar $S^{\mathrm{evid}}_{ip}$, and coverage $C_{ip}$ (by pillar and on average). As a basic check, confirm that the provider publishes: a vulnerability disclosure policy and handling process, a redress channel with stated response times, and a current model/system card describing purpose, data governance, intended uses/limits, and incident history. For research comparisons and longitudinal tracking, report the known‑only midpoint $\mathrm{AIPI}_{\text{known}}$ together with the full interval $[\mathrm{AIPI}_{\text{evid}},\mathrm{AIPI}_{\text{opt}}]$ and pillar‑level coverage.

\subsection{Uncertainty and reliability}
Uncertainty arises from coding disagreement and incomplete evidence. Each release double‑codes a stratified sample and reports inter‑rater reliability and documents sampling decisions \citep{Creswell2013,krippendorff2018content}. Sensitivity analysis varies indicator inclusion and performs leave‑one‑pillar‑out checks; rank stability is summarized using Kendall’s $\tau$ between alternative score vectors \citep{Creswell2013}. Release notes publish coverage distributions and rank intervals.

\section{Indicators and coding guide}
Indicators are designed to be \emph{observable, verifiable, and consequential}. Each code must be supported by a link to a public artifact (policy, governance record, model/system card, audit report, standard conformance statement) or a vetted external evaluation. Detailed rubrics and edge cases are available in the public codebook on GitHub:

\url{https://github.com/rsdmu/aipi-pluralism-index}

Table~\ref{tab:scoring-rubric} summarizes the generic ordinal mapping used when indicators admit partial credit.

\begin{table}[t]
\centering
\caption{Generic scoring rubric for ordinal indicators}
\label{tab:scoring-rubric}
\renewcommand{\arraystretch}{1.2}
\setlength{\tabcolsep}{6pt}
\begin{tabularx}{\linewidth}{
  >{\raggedleft\arraybackslash}p{0.10\linewidth}
  >{\raggedright\arraybackslash}p{0.28\linewidth}
  X}
\toprule
\textbf{Score} & \textbf{Rating} & \textbf{Evidence criteria / description} \\
\midrule
1.0 & Verified strong practice &
Public artifact with scope, recency, and governance remit documented; independently reproducible or externally verified. \\
0.5 & Partial evidence &
Some evidence or limited scope; unclear coverage or recency; partial implementation. \\
0.0 & No evidence &
No verifiable artifact, or claim inconsistent with protocol. \\
\bottomrule
\end{tabularx}
\end{table}

\paragraph{Participatory governance.}
This pillar codes whether affected stakeholders can provide input that is documented, deliberated, and linked to resulting actions. Qualifying evidence consists of (i) multi‑stakeholder advisory bodies with a public charter defining remit, membership selection, conflict‑of‑interest handling, and publication of agendas and minutes; (ii) public consultations or request‑for‑comment processes that publish submissions in aggregate and a disposition matrix mapping proposals to accept/reject/rationale; (iii) issue trackers or proposal forums open to external contributors where each proposal has a dated, public disposition and, when adopted, a pointer to the corresponding change log or release note; (iv) external community red teaming with a published scope, methodology, consent and safety procedures, and a post‑exercise report that identifies specific changes or risk acceptances; and (v) commitments to avoid undue concentration of power that are evidenced by implemented access or governance rules (e.g., non‑exclusive licensing terms, neutral API access criteria, data export or interoperability guarantees, limits on exclusive partnerships) rather than aspirational statements \citep{Arnstein1969,Bohman2000,ogp2024_threeways,fung2006}. Because AIPI is a global index, we also code the geographic scope of participatory mechanisms when evidence permits. For advisory bodies, public consultations, and community red teaming, full credit (1.0) requires evidence of participation that is multi-region (e.g., a roster or contributor list spanning multiple world regions, or an open process accessible internationally, including across languages where relevant). Partial credit (0.5) is assigned when a mechanism exists but participation appears limited to a single country/region, or when the geographic composition is not disclosed; no credit (0.0) is assigned when no verifiable mechanism exists.

Credit is assigned to artifacts that are publicly accessible at or before the release cutoff and that show at least one completed feedback cycle (input, deliberation, decision, implementation or reasoned rejection). Ad hoc surveys without published dispositions, private advisory arrangements without a public mandate, and announcements without links to decisions do not qualify.

\paragraph{Inclusivity and diversity.}
This pillar codes whether design, staffing, evaluation, and access reflect populations likely to be affected. Evidence includes (i) representation statistics for the teams and leadership responsible for the system, reported with category definitions, denominators, and date stamps, where such statistics are legally publishable; (ii) structured involvement of underrepresented groups in requirements gathering, prototyping, and user research, with documentation of recruitment criteria, compensation, and how findings altered specifications or defaults; (iii) language coverage for the flagship user interface, documentation, and support channels, reported as a list of supported languages and, where available, quality or coverage tests \citep{Nekoto2020}; (iv) accessibility conformance of user‑facing interfaces with scope, testing method, and exceptions documented \citep{Goggin2019}; (v) dataset documentation that reports sampling frames, known gaps, and the relationship between dataset composition and intended populations; and (vi) disaggregated evaluation results that report performance across relevant subgroups or locales, including languages beyond high‑resource settings \citep{cnrs_bloom2022}. Partnerships outside high‑income markets count when they involve co‑production rather than one‑off donations. Statements of commitment without accompanying statistics, methods, or changes do not receive credit.

\paragraph{Transparency.}
This pillar codes whether the producer discloses the information needed to understand purpose, provenance, constraints, and known limitations. Qualifying artifacts are model or system cards and datasheets that, taken together, provide: stated objectives and intended uses; governance and update cadence; training data sources and licenses or access constraints (at the level of datasets, categories, or collection pipelines as applicable); fine‑tuning and reinforcement signals; evaluation protocols and metrics, including subgroup performance where measured; documented limitations, hazards, and unsafe use cases; deployment safeguards; and incident or change histories with dates \citep{Mitchell2019,gebru2021}. Additional credit is assigned for technical or governance reports that document internal review processes with roles, decision criteria, and outcomes, and for responsiveness to independent inquiries, evidenced by a public contact channel, a stated process for requests, and example responses or Q\&A published during the release window \citep{bommasani2023fmti,Hagendorff2020}. Marketing materials without underlying artifacts, non‑persistent links, and self‑descriptions that cannot be reconciled with released systems are out of scope.

\paragraph{Accountability.}
This pillar codes whether mechanisms exist to detect, report, and remediate harms and defects, and whether external scrutiny has traction. Evidence includes (i) documented risk assessment and incident response processes that specify roles, severity classification, escalation paths, post‑incident analysis, and publication practices; (ii) a vulnerability disclosure process that states in‑scope assets, how to report, safe‑harbor language, and researcher acknowledgement; (iii) a vulnerability handling process covering triage, tracking, remediation, and communication; (iv) independent audits that name the auditor, scope, standard or methodology, sampling or testing approach, findings, and remediation timelines, with public summaries of results; (v) user redress and appeal channels with eligibility, steps, evidence requirements, and time‑bound service levels, plus aggregate statistics where published; (vi) whistleblowing channels with independence from line management and non‑retaliation statements; and (vii) participation in relevant standards processes or registries when applicable \citep{Lepri2018,Leslie2019,Reisman2018}. Additional credit is awarded when incident reports or postmortems contribute to public learning repositories, with links from incidents to mitigations. Legal compliance statements and unaudited self‑attestations without verifiable artifacts do not receive credit.

\section{Cohort, eligibility, discovery, and timelines}
\label{sec:cohort}

\textbf{Eligibility.} We included a producer if, by the cutoff date, it met all of the following: (i) develops or maintains a general‑purpose foundation model or a generative AI service that is widely available; (ii) controls releases, governance, or deployment guardrails; and (iii) has at least one system family or API that is publicly identifiable. We applied the same criteria to government consortia and non‑profit organizations.

\textbf{Provider discovery.} We identified candidate producers using four sources: (1) Google search and news coverage; (2) academic and industry reports; (3) open‑source platforms and model registries; and (4) references in standards documents, audits, and government procurement records.\footnote{We did not collect any personal data.}

\textbf{Timeline and data collection.} We set the eligibility cutoff at 30~September~2025. We compiled the dataset over four months, from early~May~2025 to late~September~2025. During this period, we collected and timestamped evidence for each producer.

\section{Data sources and pipeline}
The pipeline integrates three streams. First, structured web and repository search traverses producer sites, documentation hubs, transparency reports, model and dataset cards, and registries; enumeration uses Google search and curated news articles to seed candidate providers, then applies de‑duplication and eligibility filters (Sec.~\ref{sec:cohort}). Reproducibility metadata include query templates, URLs, timestamps, and retrieval logs. Second, external independent evaluations and benchmarks are incorporated with explicit citation, for example large‑scale capability benchmark syntheses and public safety assessments \citep{RAY2023,bengio2024safety}. Third, expert interviews with researchers, journalists, and civil society practitioners provide context for interpreting artifacts and identifying missing documentation; interviews obtain informed consent and exclude sensitive data. Evidence links are checked for accessibility and archived when permitted. The evidence graph ties producers to indicators, sources, and coders; adjudication logs are public. All builds are deterministic and versioned.

\section{Validation}
Validation proceeds on three axes. Content validity is established by triangulating the taxonomy against recognized frameworks that articulate participation, inclusion, transparency, and accountability \citep{euai2024,Tabassi2023,Goggin2019,Jobin2019}. Construct validity is probed by testing expected correlations: transparency pillar scores should correlate with external transparency syntheses without complete overlap, inclusivity indicators on language and accessibility should correlate with independent coverage measures, and accountability indicators should align with the presence of disclosure programs and audits. Reliability is addressed through coder training, a detailed codebook with exemplars and counter‑examples, double coding of a stratified sample each release, and public adjudication. We report inter‑rater agreement and document sampling, and target acceptable thresholds for publication \citep{Creswell2013}. Sensitivity analysis varies pillar inclusion and indicator weights and reports rank stability (e.g., Kendall’s $\tau$). To bound uncertainty from missing evidence we publish the evidence and known‑only scores plus coverage; for procurement use we recommend minimum thresholds using the lower‑bound evidence score.

\section{Results}
\begin{figure}[t]
  \centering
  \includegraphics[width=\linewidth]{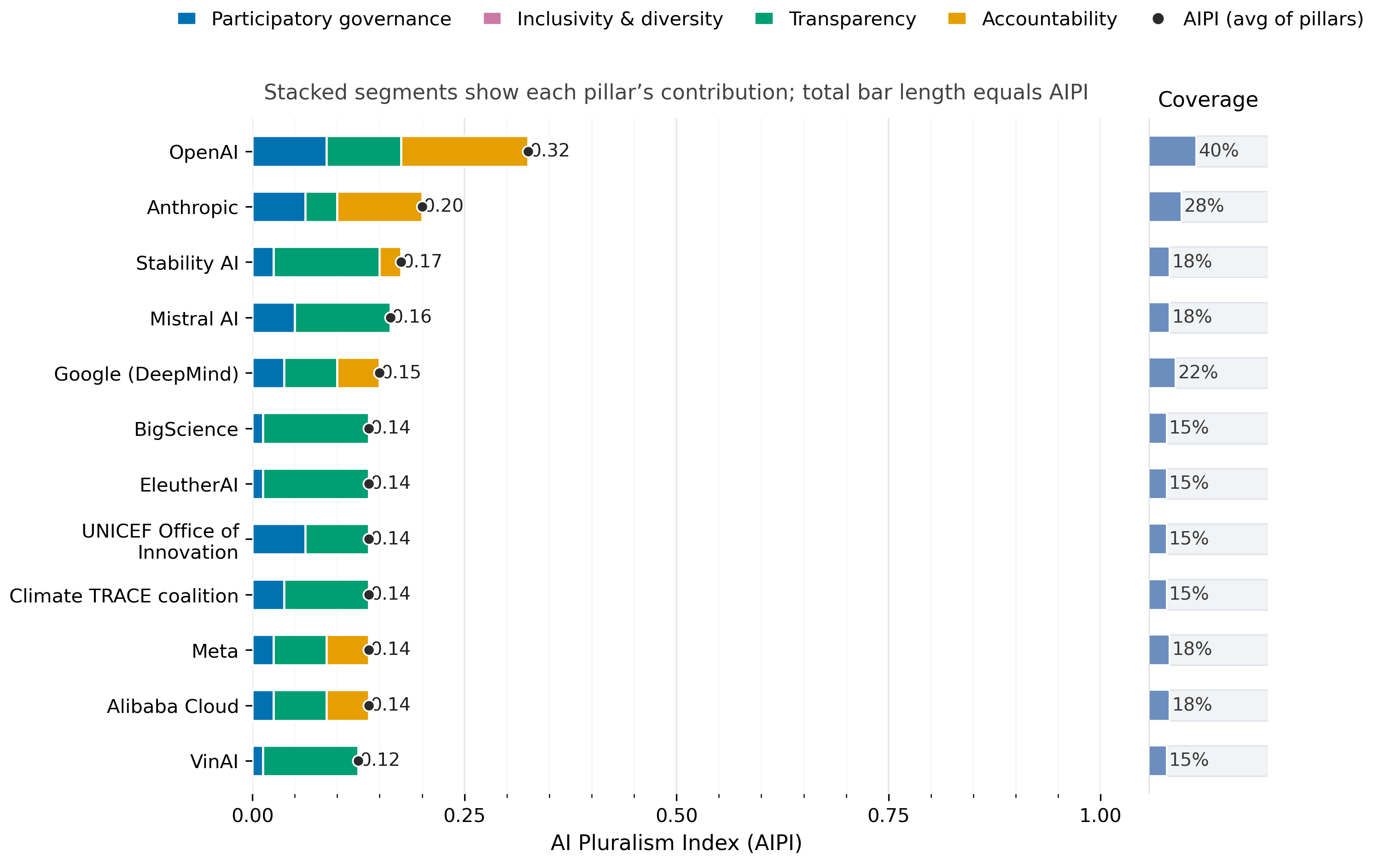}
  \caption{Provider results, known‑only treatment. Stacked segments show each pillar’s contribution (each pillar divided by four); total bar length equals the AIPI value on a 0--1 scale. The right column reports coverage, the share of indicators with evidence‑coded values.}
  \label{fig:aipi-providers-known}
\end{figure}

Figure~\ref{fig:aipi-providers-known} summarizes provider scores under the known‑only treatment for an initial cohort. Bars show each pillar’s contribution to the overall AIPI (each pillar divided by four). The right column reports coverage, the share of indicators with evidence‑coded values.

In this pilot cohort, OpenAI scores $0.32$ with the highest coverage ($\approx 40\%$). Anthropic follows at $0.20$ (coverage $\approx 28\%$). Stability AI, Mistral AI, and Google DeepMind cluster between $0.15$ and $0.17$, while BigScience, EleutherAI, UNICEF Office of Innovation, Climate TRACE, Meta, and Alibaba Cloud are near $0.14$; VinAI registers $0.12$. Coverage ranges from roughly $15\%$ to $40\%$ with a median near $18\%$. Within‑provider variation across pillars is informative: some producers obtain most credit from transparency while lagging in participatory governance or accountability, suggesting opportunities for targeted improvement. Because known‑only scores exclude missing indicators, we report the evidence scores alongside coverage in the release dataset; the difference between the two reflects documentation gaps rather than practice alone.

\section{Governance, implementation, and releases}
AIPI is implemented as a public repository with deterministic builds and archived artifacts. Anyone can file an issue to challenge a code, propose an indicator change, or contribute evidence. Issues are triaged by volunteer reviewers; disagreements are logged and escalated to a multi‑stakeholder maintainer council spanning research, civil society, industry, and regions. Each release is versioned and archived with a change log. The website provides system and provider profiles, an API, and download links for evidence tables and scores. Governance adheres to the same pluralistic principles that the instrument evaluates and aligns with open practices in accessibility and security communities \citep{Goggin2019,Lepri2018}.

We operationalize this governance with three layers of decision making. First, a release team executes evidence discovery, coding, and quality assurance using a frozen codebook for that version. Second, an adjudication panel resolves disputed codes using a publicly documented standard of evidence, logging rationales and precedents so that similar cases are treated consistently across releases. Third, the maintainer council decides on indicator additions, deprecations, and major rubric changes through a transparent proposal process that includes (i) an open call for change requests, (ii) a public comment window, and (iii) recorded decisions with an explicit migration plan for longitudinal comparability.

To address concentration and geographic skew in participation, the maintainer council is constituted to include representation across stakeholder categories and regions (research, civil society, industry, labor, and policy), with time-bounded terms and conflict-of-interest disclosures. Members recuse themselves from decisions involving their own organizations. Releases follow a predictable cadence with an evidence freeze date, double-coding of a stratified sample, and a short pre-release audit period during which external contributors can contest evidence links or propose missing artifacts. Final releases are archived, and we publish a changelog that distinguishes schema changes (which affect comparability) from evidence updates (which affect scores).

\section{Recommendations for practice and policy}
For producers, a practical path to improving AIPI is to publish verifiable artifacts that reflect real processes, and to ensure that those processes include affected stakeholders across deployment contexts. Model and system cards should disclose data governance, known limitations, intended uses, subgroup performance, and update histories. Vulnerability disclosure processes should accept reports from good-faith researchers and specify service levels; vulnerability handling should name a PSIRT (Product Security Incident Response Team) function and lay out triage and remediation steps. Public consultations should summarize input and publish a disposition matrix that links proposals to decisions and downstream changes, rather than collecting comments without response. Where systems are deployed internationally, participatory mechanisms should be accessible across regions, for example via remote participation, multilingual materials, and geographically diverse advisory membership.

For procurers, AIPI can be used as a practical checklist rather than a headline ranking. Reference the indicator list directly in requests for proposals; require minimum evidence scores and explicit coverage thresholds; and ask vendors to supply the primary artifacts (not summaries) referenced by their scores. Use pillar-level scores to tailor due-diligence: for low participatory-governance scores, require a roadmap for consultation and redress; for low transparency scores, require current system cards and change logs; for low accountability scores, require incident response and audit documentation.

For regulators and standards bodies, several indicators map onto emerging governance requirements, including documentation, risk management, incident reporting, accessibility, and redress. AIPI can support supervision by translating these requirements into auditable evidence checks and by highlighting where documentation is absent, stale, or geographically narrow. Regulators can also use AIPI’s coverage reporting to prioritize oversight, focusing audits where low scores are paired with low evidence coverage.

For researchers and civil society, AIPI’s evidence graph enables systematic scrutiny of claims and targeted advocacy that rewards concrete improvements. We encourage third-party replications and ``shadow scores'' for contested providers, and we treat well-documented replications as admissible evidence in subsequent releases.
\section{Limitations}
AIPI relies on observable artifacts and external evaluations, so it may undercount non‑public internal practices. It could be gamed if producers publish documentation without corresponding practice. We mitigate by requiring time‑stamped artifacts, awarding credit for external audits and incident disclosures that create real accountability, and publishing coverage to make documentation gaps visible. The index inherits biases from the availability of English‑language documentation; we prioritize multilingual evidence and weight indicators that reward accessibility and language coverage. We do not measure capability or risk directly, and the index should be used alongside capability, safety, and domain‑specific evaluations \citep{RAY2023,BommasaniEtAl2021,bengio2024safety}. Some sources in the ecosystem are journalistic or blog posts; we treat these as context and anchor core claims to primary or official texts where possible. 

A further limitation concerns scope and survivorship bias in community projects. Because the unit of analysis is the producer and, where applicable, system families offered as ongoing services, many participatory, community‑led “plural‑AI” initiatives—often experimental, one‑off, or currently inactive—fall outside the present cohort and are therefore underrepresented. This omission risks overweighting well‑resourced providers and understating pluralistic innovation emerging from research collectives, civic labs, hackathons, and local co‑design efforts. To mitigate this, future releases will pilot a community‑maintained registry and evidence track for plural‑AI projects, including non‑service and archived initiatives that meet minimal provenance requirements (public or archived artifacts with dates, licensing, and an explicit activity status). Crowdsourced nominations and review, integrated through our adjudication and versioning processes, would sustain a living record that expands coverage over time and makes plural models more discoverable to users and procurers. Results from this track will be reported separately to avoid conflating prototypes and time‑bounded pilots with production systems.

Finally, the cohort is expanding; early release comparisons should be interpreted as a baseline rather than a definitive ranking.

\section{Theory of change}
By centering participation, inclusion, transparency, and accountability—and by embedding those principles in AIPI’s governance—the instrument seeks to align reputational and procurement incentives with democratic values. The combination of public evidence, coverage reporting, and versioned builds enables regulators and procurers to reference specific indicators in policy or contracts. Capability and safety benchmarks address performance and risk; AIPI addresses whether governance is pluralistic and answerable. Together, they provide a broader basis for responsible selection and oversight.

\section{Future work}
\label{sec:future}
Next releases will (i) add multilingual evidence discovery protocols, including targeted search in major languages and structured prompts for local documentation; (ii) reward third‑party local verifications, privileging evidence from independent community groups and regional agencies; and (iii) document regional coverage explicitly by tagging artifacts with geography and publishing pillar‑level coverage intervals by region. We will release partial‑dependence plots of AIPI versus coverage and report where rank flips occur; policy guidance will recommend $\mathrm{AIPI}_{\text{evid}}$ for procurement and $\mathrm{AIPI}_{\text{known}}$ with intervals for comparative research. We will also extend correlations against transparency syntheses, capability facets, and safety reports to the growing cohort and publish bootstrapped confidence intervals.

\section{Conclusion}
Pluralistic AI requires structures that allow many voices to shape technical and governance choices, and mechanisms that render those structures visible and answerable. The AI Pluralism Index provides a principled, auditable, and adaptable way to measure progress toward that goal. We will publish regular releases with evidence graphs, adjudication logs, reliability statistics, and sensitivity analyses. The first public version is available at:

Index: \url{https://aipluralism.wiki/}\\
Code: \url{https://github.com/rsdmu/aipi-pluralism-index}

These resources provide open access to the framework, documentation, and implementation, enabling replication, scrutiny, and continued community contribution.

\paragraph{Ethics statement.}
This work analyzed publicly available artifacts and conducted expert interviews under informed consent, without collecting personal or sensitive information. Interview protocols, coding instructions, de‑identified materials, and analysis scripts will be released in the project repository upon publication.

\bibliographystyle{plainnat}
\bibliography{references}

\appendix

\section{Indicator examples and coding excerpts}
Selected examples (full codebook in the repository):

\begin{longtable}{p{3.0cm}p{11.0cm}}
\toprule
\textbf{Pillar} & \textbf{Indicator and coding excerpt} \\\midrule
Participatory governance &
External red teaming or community audits with published scope, methodology, and outcomes. Code $0$ if no external engagement; $0.5$ for ad hoc events, single-region engagement, or limited documentation; $1$ for recurring exercises with multi-region participation, full reports, and evidence of resulting changes. \\\addlinespace
Inclusivity \& diversity &
Multilingual access and accessibility features in flagship offerings benchmarked to clear scope and verification. Code $0$ if monolingual and inaccessible; $0.5$ for limited support; $1$ for systematic coverage including underrepresented languages, evidenced by independent evaluation. \\\addlinespace
Transparency &
Model and system cards that describe purpose, data sources, governance, intended uses, limitations, and subgroup performance. Code $0$ if absent; $0.5$ if partial; $1$ if comprehensive, current, and independently verified. \\\addlinespace
Accountability &
Redress and appeal mechanisms with time‑bound service levels and published incident statistics; presence of vulnerability disclosure and handling processes with public documentation. Code $0$ if none; $0.5$ for basic contact channels; $1$ for documented, audited processes with statistics and outcomes. \\\bottomrule
\end{longtable}

\end{document}